\documentclass{article}
\usepackage{leaplab}

\usepackage[colorlinks,linkcolor=red,citecolor=blue,urlcolor=blue]{hyperref}
\usepackage{url}
\usepackage{placeins}
\usepackage[table,x11names]{xcolor}
\usepackage[utf8]{inputenc}
\usepackage[T1]{fontenc}
\usepackage{url}
\usepackage{booktabs}
\usepackage{amsfonts}
\usepackage{wrapfig}
\usepackage{subcaption}
\usepackage{amsmath}
\usepackage{float} 
\usepackage{amssymb}
\usepackage{amsthm}
\usepackage{bm}
\usepackage{nicefrac}
\usepackage{wrapfig}
\usepackage{microtype}
\usepackage{graphicx}
\usepackage{caption}
\usepackage{subcaption}
\usepackage{algorithm}
\usepackage{algorithmic}
\usepackage{multirow}
\usepackage[most]{tcolorbox}
\usepackage{enumerate}
\usepackage{pifont}
\usepackage{amsthm}

\usepackage{fontawesome5}
\usepackage{enumitem}
\usepackage{marvosym}
\definecolor{cplblue}{RGB}{239,245,252}
\definecolor{baselinegray}{RGB}{245,247,250}

\usepackage{soul}
\usepackage{multicol}

\theoremstyle{remark}

\definecolor{lightgrey}{RGB}{247, 247, 247}
\newenvironment{leapabstract}{
  \begin{tcolorbox}[
    colback=lightgrey,
    colframe=white,
    boxrule=0pt,
    arc=10pt,
    left=16pt,
    right=16pt,
    top=12pt,
    bottom=12pt,
    width=\textwidth,
    enlarge left by=0mm,
    before skip=10pt,
    after skip=10pt
  ]
}{
  \end{tcolorbox}
}

\makeatletter
\def\icmldate#1{\gdef\@icmldate{#1}}
\icmldate{\today}
\makeatother

\makeatletter
\fancypagestyle{fancytitlepage}{
  \fancyhead{}
  \lhead{\includegraphics[height=1.5cm]{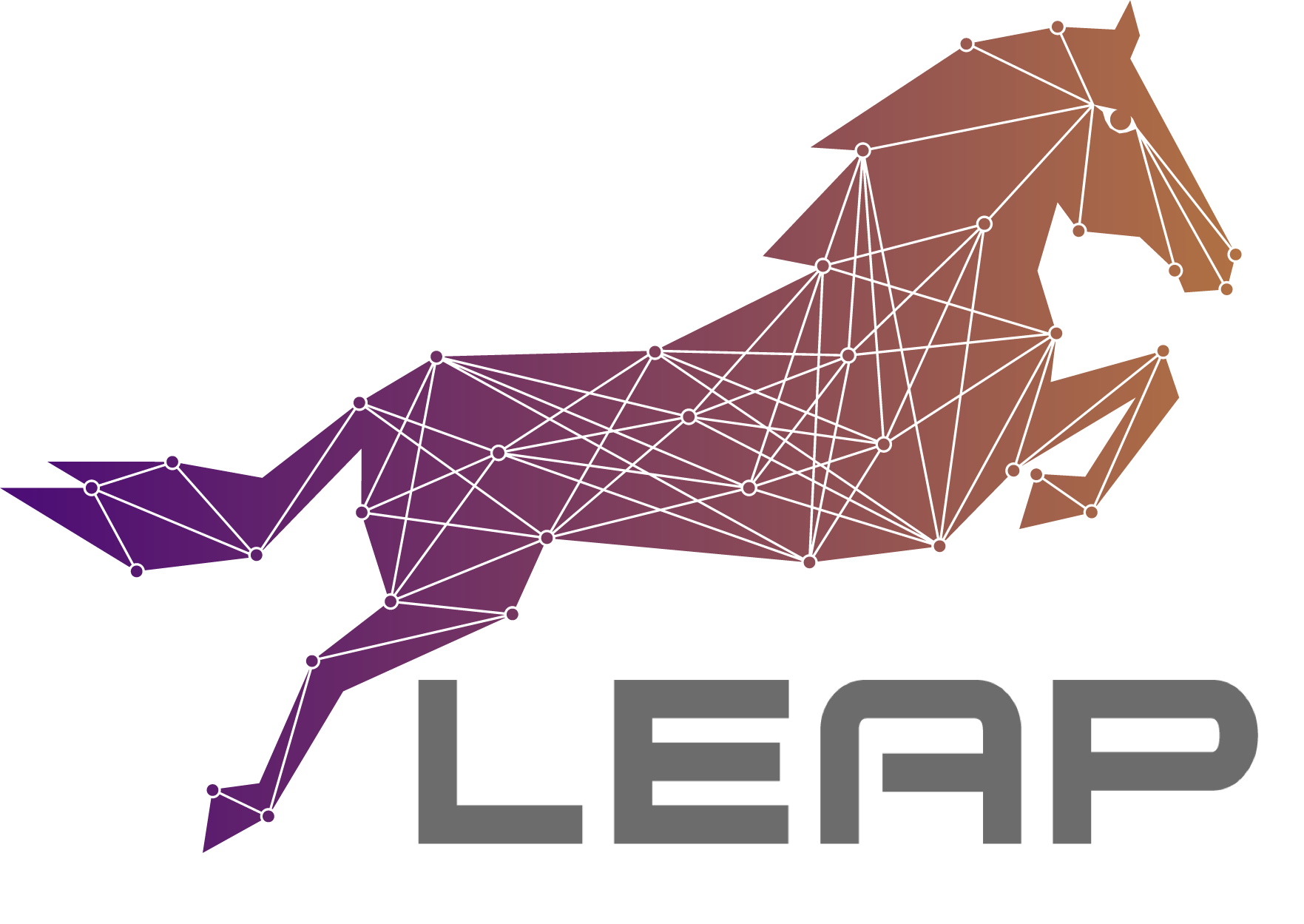}}
  \rhead{\it \@icmldate}
  \cfoot{}
}
\makeatother

\icmltitlerunning{Do We Really Need KL Divergence for On-Policy Distillation of Large Language Models?}

\begin{document}

\icmltitle{Do We Really Need KL Divergence for On-Policy Distillation of Large Language Models?}

\begin{icmlauthorlist}
\mbox{Wenze Lin$^{\,1,2\,*\,\dagger}$},
\mbox{Jiyuan Long$^{\,2\,*}$},
\mbox{Jiale Zhao$^{\,1,3\,*}$},
\mbox{Shenzhi Wang$^{\,1}$},
\mbox{Xitai Jiang$^{\,1,2}$},
\mbox{Ce Luo$^{\,4}$},
\mbox{Rui Lan$^{\,5}$},
\mbox{Qianli Ma$^{\,1,2}$},
\mbox{Fukang Wen$^{\,1,2}$},
\mbox{Hui Wu$^{\,6}$},
\mbox{Liyuan Chen$^{\,6}$},
\mbox{Shuoling Liu$^{\,6}$},
\mbox{Jiangpeng Yan$^{\,6}$},
and \mbox{Gao Huang$^{\,1\,\textrm{\Letter}}$}
\end{icmlauthorlist}

$^{1\,}$LeapLab, Tsinghua University \quad
$^{2\,}$Qiuzhen College, Tsinghua University \quad
$^{3\,}$Beihang University \quad
$^{4\,}$National University of Singapore \quad
$^{5\,}$The Chinese University of Hong Kong \\
$^{6\,}$E Fund Management Co., Ltd. \quad

$^{*}$ Equal Contribution \quad $^{\dagger}$ Project Lead \quad $^{\textrm{\Letter}}$ Corresponding Author

\icmlcorrespondingauthor{linwz25@mails.tsinghua.edu.cn, gaohuang@tsinghua.edu.cn}{}

\vskip .3in

\printNotice{}

\begin{leapabstract}
Since the advent of knowledge distillation, KL divergence has been the standard loss in distillation. Recently, on-policy distillation (OPD) has emerged as an efficient post-training paradigm for LLMs. As a distillation method, OPD naturally inherits KL divergence as its standard loss. However, in this work, we find that KL divergence may not be necessary for OPD. We show that simply preserving the update direction is sufficient for effective OPD. As long as the update direction is toward the teacher, OPD works. More precisely, it is not the direction of every token, but the direction of a small subset of tokens where the teacher and student disagree strongly. We first show that simply assigning a reward of \(+1\) to tokens where the teacher probability is higher than the student probability and \(-1\) where it is lower, which merely encourages updates toward the teacher, reproduces almost the same training mode as OPD with reverse KL. We further show that only the direction of a small subset of tokens with large teacher-student disagreement is critical, and training works as long as their update direction is toward the teacher, even if other tokens are pulled away from the teacher. And as an application of these findings, we introduce Consensus Multi-Teacher On-Policy Distillation (C-MOPD) to improve Multi-Teacher On-Policy Distillation (MOPD). Unlike MOPD, which routes each sample to a single teacher and may cause capability conflicts across domains, C-MOPD lets every sample be supervised by all teachers. Experiments show that C-MOPD consistently outperforms MOPD on both math and code benchmarks. Our code is available at \url{https://github.com/LeapLabTHU/KL-Free-OPD}.
\end{leapabstract}

\section{Introduction}\label{sec:introduction}
Since the advent of knowledge distillation~\citep{hinton2015distilling}, KL divergence has been the default loss in distillation, as it allows the student to match the teacher's soft targets and thereby capture the dark knowledge encoded in the teacher's output distribution. Recently, on-policy distillation (OPD)~\citep{xu2026deepseek,xiao2026mimo,yang2025qwen3,zeng2026glm} has emerged as an efficient post-training paradigm for LLMs. While classical distillation trains the student on a fixed dataset, OPD aligns the student with the teacher on the student's own rollouts. As a distillation method, OPD naturally inherits KL divergence as its standard loss.
 Minimizing the reverse KL divergence between the student and the teacher on the student's own rollouts preserves the on-policy property and achieves strong performance~\citep{song2026survey,wu2025rethinking,zhang2026fast,zhao2026decoupling}. In this work, however, we find that KL divergence may not be necessary for OPD. We show that simply preserving the update direction, without its magnitude, is sufficient for effective OPD. As long as the update direction is toward the teacher, OPD works. More precisely, it is not the direction of every token, but the direction of a small subset of tokens where the teacher and student disagree strongly. We first show that simply assigning a reward of \(+1\) to tokens where the teacher probability is higher than the student probability, and \(-1\) where the teacher probability is lower than the student probability, which merely encourages updates toward the teacher, reproduces almost the same training mode as OPD with reverse KL and achieves comparable or better results on code and math benchmarks. We further show that it is not the direction of every token that matters, but only that of a small critical subset of tokens with large teacher-student disagreement: OPD works as long as their update direction is toward the teacher, and OPD fails as long as their update direction is away from the teacher. In our experiments, we find that keeping only high-disagreement tokens, which in some cases account for less than \(2\%\) of all tokens, and ensuring their update direction is toward the teacher reproduces OPD; once their direction is reversed, OPD fails to train. We also find that the low-disagreement tokens, which constitute the vast majority of tokens, can be updated away from the teacher without severely affecting OPD. These findings challenge the prevailing assumption that KL divergence is essential for on-policy distillation, and suggest that the key ingredient is not distribution matching but merely directional updates on a small subset of high-disagreement tokens.

As an application of these findings, we propose Consensus Multi-Teacher On-Policy Distillation (C-MOPD) to improve Multi-Teacher On-Policy Distillation (MOPD)~\citep{ma2026mopd,blakeman2025nvidia,team2026kimi}. In MOPD, for each sample from the student's own rollouts, the student computes the reverse KL divergence against a domain-specific teacher and updates accordingly. As the number of samples increases, the student integrates the capabilities of different teachers. But a drawback is that updating on samples from one domain may degrade the model's capability in another domain. In C-MOPD, every update is guided by all teachers: when all teachers agree on the update direction, we pull the student toward all of them; when the teachers conflict, we fall back to the domain-specific teacher and adopt its direction only if it does not strongly conflict with the other teachers, and otherwise we assign a zero reward. This design is motivated by our earlier finding that in OPD, tokens with small teacher-student disagreement can be updated away from the teacher without hurting performance. In this way, C-MOPD ensures that each update either pulls toward all teachers, or toward the domain-specific teacher without hurting the capabilities associated with the other teachers' directions. Experiments show that C-MOPD consistently outperforms MOPD.


Our contributions can be summarized as below:
\begin{itemize}
    \item We find that simply ensuring the update direction is toward the teacher is sufficient for OPD, questioning whether KL divergence is necessary for OPD.
    \item We further find that only the direction of tokens with large teacher-student disagreement is critical. In some cases, keeping only less than \(2\%\) of the tokens while ensuring their direction is toward the teacher makes OPD work. 
    \item We propose Consensus Multi-Teacher On-Policy Distillation (C-MOPD) to improve MOPD, which guides every token with all teachers, consistently outperforming MOPD.
\end{itemize}

\section{Related Work}\label{sec:related_work}

\paragraph{On-Policy Distillation with Reverse KL.}
On-policy Distillation (OPD) has recently emerged as an efficient post-training paradigm that provides dense, token-level supervisory signals by distilling a teacher model's output distribution into the student~\citep{xu2026deepseek,xiao2026mimo,yang2025qwen3,zeng2026glm,agarwal2024policy,li2026rethinking}. The de facto loss for OPD is reverse KL divergence, which is mode-seeking and preserves the on-policy property~\citep{zhao2026decoupling, agarwal2024policy,shao2026token,song2026survey,wu2025rethinking,zhang2026fast}. In this work, however, we question whether reverse KL is truly necessary for effective teacher guidance. We find that simply assigning \(+1\) to tokens where the teacher probability is higher than the student probability and \(-1\) where it is lower already reproduces almost the same training mode as OPD, with comparable or better results.

\paragraph{Multi-Teacher On-Policy Distillation.}
Multi-teacher on-policy distillation (MOPD) has become a common approach for integrating the capabilities of multiple teacher models into a single student~\citep{ma2026mopd,blakeman2025nvidia,team2026kimi,gao2026open, chen2026counteraction,sun2026d,he2026learn}. In MOPD, each sample is routed to a single domain-specific teacher, so that the student gradually acquires different teachers' expertise as it sees more samples. However, based on our observation that OPD only needs to maintain updates toward the teacher, we propose Consensus Multi-Teacher On-Policy Distillation (C-MOPD), which allows every sample to receive guidance from all teachers. This avoids the loss of one domain's capability when updating on samples from another domain.


\section{Preliminary} \label{sec:preliminary}
\paragraph{On-Policy Distillation with Reverse KL.}

On-Policy Distillation (OPD) aims to transfer the capability of a teacher policy \(\pi_T\) to a student policy \(\pi_\theta\) by aligning the student with the teacher on the student's own rollouts. Formally, given a query \(q\), the student generates an output sequence \(o \sim \pi_\theta(\cdot \mid q)\), and the default objective is to minimize the reverse KL divergence between the two policies at every token position:
\[
\min_{\theta} \ \mathbb{E}_{q, \, o \sim \pi_\theta(\cdot \mid q)} \left[ \sum_{t=1}^{|o|} D_{\mathrm{KL}}\big( \pi_\theta(\cdot \mid q, o_{<t}) \,\big\|\, \pi_T(\cdot \mid q, o_{<t}) \big) \right].
\]
Because the expectation is taken over trajectories sampled from \(\pi_\theta\) itself, the training remains on-policy and benefits from dense token-level supervision. To derive the update rule, one can differentiate the objective and obtain a policy-gradient-like expression:
\[
\nabla_\theta \mathcal{L}_{\text{OPD}}(\theta)
=
\mathbb{E}_{q, \, o \sim \pi_\theta(\cdot \mid q)}
\left[
\sum_{t=1}^{|o|}
\Big( \log \pi_T(o_t \mid q, o_{<t}) - \log \pi_\theta(o_t \mid q, o_{<t}) \Big)
\nabla_\theta \log \pi_\theta(o_t \mid q, o_{<t})
\right].
\]
This expression shows that each token receives a dense reward, which is exactly the log-ratio
\[
r_t = \log \frac{\pi_T(o_t \mid q, o_{<t})}{\pi_\theta(o_t \mid q, o_{<t})}.
\]

\paragraph{Multi-Teacher On-Policy Distillation.}
Multi-Teacher On-Policy Distillation (MOPD) extends OPD to multiple domain-specific teachers. Let \(\{\pi_{T_1}, \dots, \pi_{T_K}\}\) denote \(K\) teachers, each specialized in a different domain. For each query \(q\), a routing function \(d(q) \in \{1,\dots,K\}\) selects the teacher corresponding to the domain of \(q\). The student is then optimized to minimize the reverse KL divergence against the routed teacher on its own rollouts:
\[
\min_{\theta} \ \mathbb{E}_{q, \, o \sim \pi_\theta(\cdot \mid q)} \left[ \sum_{t=1}^{|o|} D_{\mathrm{KL}}\big( \pi_\theta(\cdot \mid q, o_{<t}) \,\big\|\, \pi_{T_{d(q)}}(\cdot \mid q, o_{<t}) \big) \right].
\]

\section{Teacher-Directional Updates Suffice for On-Policy Distillation}
\label{sec:BinaryOPD}
\subsection{Method}
\label{sec:BinaryOPD-method}

We replace the reverse KL objective in OPD with a simple directional reward.

Given a query \(q\), the student generates a rollout \(o \sim \pi_\theta(\cdot \mid q)\). For each token \(o_t\), we compare the teacher probability \(\pi_T(o_t \mid q, o_{<t})\) with the student probability \(\pi_\theta(o_t \mid q, o_{<t})\). We assign a reward
\[
r_t =
\begin{cases}
+1, & \text{if } \pi_T(o_t \mid q, o_{<t}) > \pi_\theta(o_t \mid q, o_{<t}), \\[2pt]
-1, & \text{if } \pi_T(o_t \mid q, o_{<t}) < \pi_\theta(o_t \mid q, o_{<t}).
\end{cases}
\]
Specifically, when the teacher probability is higher than the student probability, the \(+1\) reward increases the probability of that token in the student, thereby pulling the student toward the teacher. When the teacher probability is lower than the student probability, the \(-1\) reward decreases the probability of that token in the student, again moving the student toward the teacher. In summary, this reward merely ensures that the student updates in the direction of the teacher, without any magnitude information. Note that the case where the two probabilities are exactly equal is almost impossible in practice, so we do not include it in the formula above; in our implementation, we assign a reward of \(0\) to such tokens.

The student is then optimized with a policy gradient objective:
\[
\max_{\theta} \ 
\mathbb{E}_{q, \, o \sim \pi_\theta(\cdot \mid q)}
\left[
\sum_{t=1}^{|o|} r_t \log \pi_\theta(o_t \mid q, o_{<t})
\right].
\]
We refer to this method as \textbf{BinaryOPD}.

\subsection{Experiments}
\label{sec:bdr-experiments}

\subsubsection{Experimental Setup}
\label{sec:bdr-exp-setup}

We compare standard OPD against our BinaryOPD on both mathematical reasoning and code generation tasks. For each student-teacher pair, we train the student with either the reverse KL divergence (OPD) or the binary-directional reward (BinaryOPD) on the same dataset, using identical training hyperparameters, rollout budgets, and optimization settings.

For math, we use five student-teacher pairs covering different model families and scales. For code, we use two student-teacher pairs. The training datasets are DAPO-Math-17k~\citep{yu2026dapo}, DeepMath~\citep{he2026deepmath}, and Eurus~\citep{cui2025process}. Table~\ref{tab:experiment_setup} summarizes all pairs.

\begin{table}[h]
\centering
\caption{Student-teacher pairs and training datasets used in our experiments.}
\label{tab:experiment_setup}
\begin{tabular}{lll}
\toprule
Student & Teacher & Dataset \\
\midrule
\multicolumn{3}{l}{\textit{Math}} \\
DeepSeek-Distill-Qwen-1.5B & JustRL-1.5B & DAPO-Math-17k \\
Qwen3-1.7B-Base & Qwen3-4B-Base-RL & DAPO-Math-17k \\
Llama-3.2-3B-Instruct & GT-Llama3.2-3B-MATH & DeepMath \\
Qwen3-4B-Non-Thinking & Qwen3-4B-Non-Thinking-RL-Math & DeepMath \\
Qwen3-30B-A3B-Non-Thinking & Qwen3-30B-A3B-Instruct-2507 & DeepMath \\
\midrule
\multicolumn{3}{l}{\textit{Code}} \\
Qwen3-4B-Non-Thinking & Qwen3-4B-Non-Thinking-RL-Code & Eurus \\
DeepSeek-Distill-Qwen-1.5B & Nemotron-Research-Reasoning-1.5B & Eurus \\
\bottomrule
\end{tabular}
\end{table}

For math, we evaluate on AIME24, AIME25, AMC, MATH500~\citep{lightman2023let}, Minerva~\citep{lewkowycz2022solving}, and OlympiadBench~\citep{he2024olympiadbench}. For code, we evaluate on LiveCodeBench v6~\citep{jain2025livecodebench}, HumanEval~\citep{chen2021evaluating}, and MBPP~\citep{austin2021program}. Experimental details are provided in Appendix~\ref{app:Experimental_details}.

\subsubsection{Experimental Results}

\begin{table}[t]
\centering
\caption{Math benchmark results. All results are averaged over 8 samples (Avg@8). }
\label{tab:math-results}
\small
\setlength{\tabcolsep}{4pt}
\begin{tabular}{@{}l*{7}{c}@{}}
\toprule
\textbf{Method} & \textbf{AIME24} & \textbf{AIME25} & \textbf{AMC}
& \textbf{MATH500} & \textbf{Minerva} & \textbf{Olympiad} & \textbf{Avg} \\
\midrule
\multicolumn{8}{@{}l}{\textit{DeepSeek-Distill-Qwen-1.5B (Student) vs. JustRL-1.5B (Teacher)}} \\
\addlinespace[2pt]
\rowcolor{baselinegray}
Student & 23.7 & 17.0 & 59.7 & 80.9 & 23.7 & 41.4 & 41.0 \\
\rowcolor{baselinegray}
Teacher & 46.2 & 37.5 & 82.0 & 89.2 & 30.7 & 53.5 & 56.5 \\
\addlinespace[2pt]
\rowcolor{cplblue}
OPD & 43.8 & \textbf{32.5} & \textbf{78.8} & 88.5 & \textbf{31.7} & 52.6 & 54.7 \\
\rowcolor{cplblue}
BinaryOPD & \textbf{45.4} & 32.1 & 78.5 & \textbf{88.9} & 31.6 & 52.6 & \textbf{54.9} \\
\midrule
\multicolumn{8}{@{}l}{\textit{Qwen3-1.7B-Base (Student) vs. Qwen3-4B-Base-RL (Teacher)}} \\
\addlinespace[2pt]
\rowcolor{baselinegray}
Student & 4.1 & 1.7 & 23.2 & 48.9 & 8.9 & 17.1 & 17.3 \\
\rowcolor{baselinegray}
Teacher & 10.6 & 13.1 & 40.3 & 74.2 & 17.2 & 30.0 & 30.9 \\
\addlinespace[2pt]
\rowcolor{cplblue}
OPD & 7.5 & 3.3 & 26.4 & 57.8 & \textbf{11.4} & 20.6 & 21.2 \\
\rowcolor{cplblue}
BinaryOPD & \textbf{8.1} & \textbf{4.0} & \textbf{30.3} & \textbf{59.1} & 10.0 & \textbf{21.6} & \textbf{22.2} \\
\midrule
\multicolumn{8}{@{}l}{\textit{Llama-3.2-3B-Instruct (Student) vs. GT-Llama3.2-3B-MATH (Teacher)}} \\
\addlinespace[2pt]
\rowcolor{baselinegray}
Student & 2.9 & 0.0 & 14.8 & 34.5 & 7.7 & 7.9 & 11.3 \\
\rowcolor{baselinegray}
Teacher & 11.7 & 0.0 & 20.9 & 49.1 & 15.7 & 15.4 & 18.8 \\
\addlinespace[2pt]
\rowcolor{cplblue}
OPD & \textbf{12.5} & \textbf{0.4} & \textbf{22.7} & 46.0 & 13.3 & 14.3 & \textbf{18.2} \\
\rowcolor{cplblue}
BinaryOPD & 9.2 & 0.0 & 21.7 & \textbf{47.2} & \textbf{14.8}& \textbf{14.6} & 17.9 \\
\rowcolor{cplblue}
\midrule
\multicolumn{8}{@{}l}{\textit{Qwen3-4B-Non-Thinking (Student) vs. Qwen3-4B-Non-Thinking-RL-Math (Teacher)}} \\
\addlinespace[2pt]
\rowcolor{baselinegray}
Student & 25.0 & 16.7 & 58.4 & 81.3 & 27.7 & 42.9 & 42.0 \\
\rowcolor{baselinegray}
Teacher & 61.2 & 56.3 & 88.0 & 93.2 & 36.7 & 60.5 & 66.0 \\
\addlinespace[2pt]
\rowcolor{cplblue}
OPD & 60.4 & 51.7 & 85.7 & \textbf{93.0} & 34.5 & 59.2 & 64.1 \\
\rowcolor{cplblue}
BinaryOPD & \textbf{61.3} & \textbf{55.0} & \textbf{85.7} & 92.8 & \textbf{35.9} & \textbf{59.8} & \textbf{65.1} \\
\midrule
\multicolumn{8}{@{}l}{\textit{Qwen3-30B-A3B-Non-Thinking (Student) vs. Qwen3-30B-A3B-Instruct-2507 (Teacher)}} \\
\addlinespace[2pt]
\rowcolor{baselinegray}
Student &33.3 &20.0 &65.2 &85.9 & 32.8&51.2 &48.1 \\
\rowcolor{baselinegray}
Teacher &74.1 & 62.1& 90.2& 95.7& 34.3& 66.8& 70.5\\
\addlinespace[2pt]
\rowcolor{cplblue}
OPD &46.7 &\textbf{31.3 }& 75.3& \textbf{89.6}&33.4 & 52.7& 54.8\\
\rowcolor{cplblue}
BinaryOPD &\textbf{47.5 }& 30.8 & \textbf{76.1} & 89.4 & \textbf{34.0} & \textbf{53.1} & \textbf{55.2}\\
\bottomrule
\end{tabular}
\end{table}

\begin{table}[t]
\centering
\caption{Code benchmark results. All results are averaged over 8 samples (Avg@8).}
\label{tab:code-results}
\small
\setlength{\tabcolsep}{4pt}
\begin{tabular}{@{}l*{4}{c}@{}}
\toprule
\textbf{Method} & \textbf{LiveCodeBench v6} & \textbf{HumanEval} & \textbf{MBPP} & \textbf{Avg} \\
\midrule
\multicolumn{5}{@{}l}{\textit{Qwen3-4B-Non-Thinking (Student) vs. Qwen3-4B-Non-Thinking-RL-Code (Teacher)}} \\
\addlinespace[2pt]
\rowcolor{baselinegray}
Student & 25.1 & 80.8 & 48.7 & 51.5 \\
\rowcolor{baselinegray}
Teacher & 36.0 & 90.6 & 56.2 & 60.9 \\
\addlinespace[2pt]
\rowcolor{cplblue}
OPD & \textbf{27.9} & 85.4 & \textbf{53.0} & \textbf{55.4} \\
\rowcolor{cplblue}
BinaryOPD & 27.8 & \textbf{85.7} & 52.1 & 55.2 \\
\midrule
\multicolumn{5}{@{}l}{\textit{DeepSeek-Distill-Qwen-1.5B (Student) vs. Nemotron-Research-Reasoning-1.5B (Teacher)}} \\
\addlinespace[2pt]
\rowcolor{baselinegray}
Student & 18.0 & 66.1 & 40.7 & 41.6 \\
\rowcolor{baselinegray}
Teacher & 28.6 & 80.0 & 47.2 & 51.9 \\
\addlinespace[2pt]
\rowcolor{cplblue}
OPD & \textbf{27.7} & 79.6 & 45.6 & \textbf{51.0} \\
\rowcolor{cplblue}
BinaryOPD & 27.6 & \textbf{78.1} & \textbf{46.1} & 50.6 \\
\bottomrule
\end{tabular}
\end{table}

\begin{figure}[t]
\centering
\includegraphics[width=\textwidth]{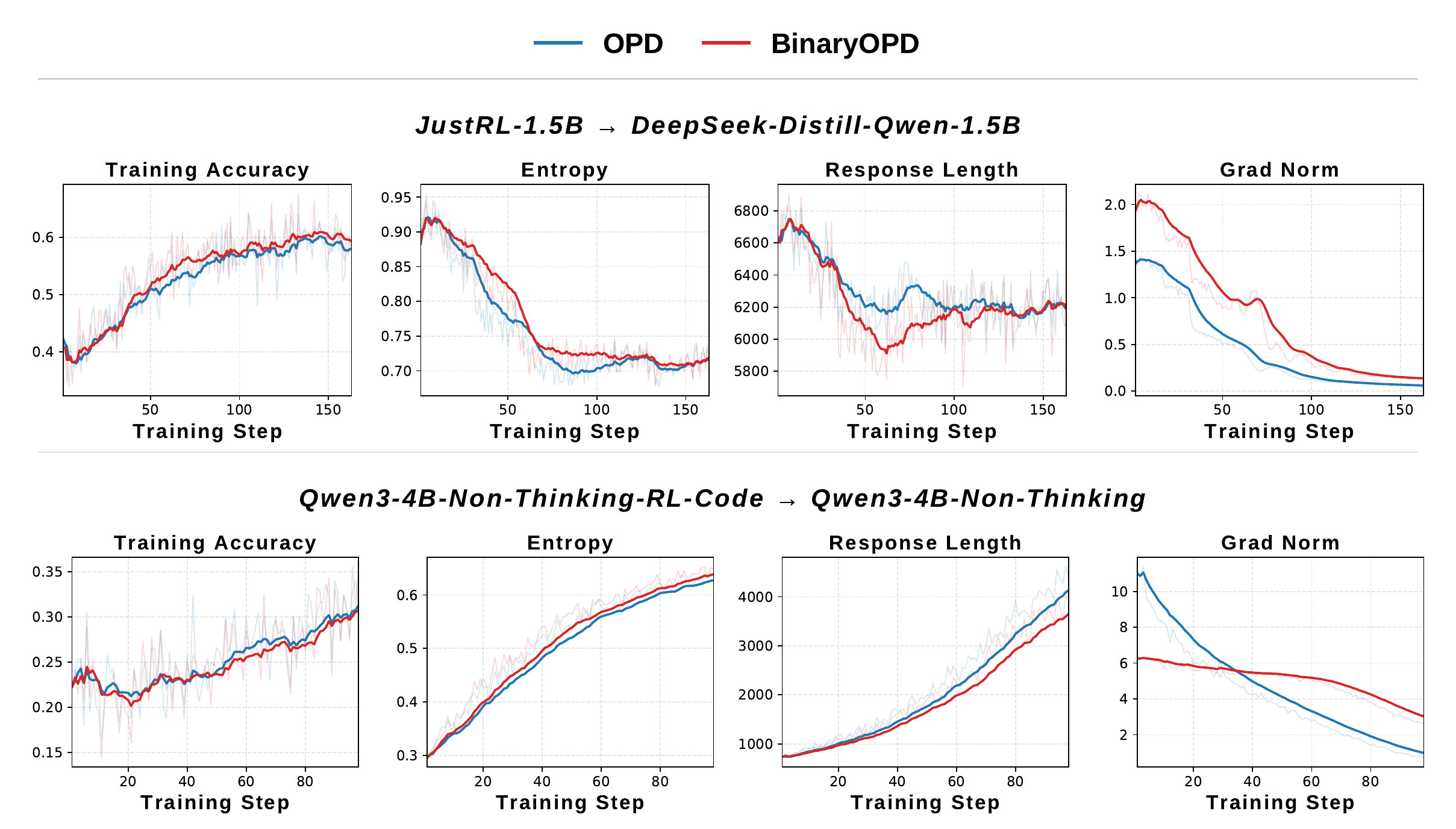}
\caption{Training dynamics of OPD and BinaryOPD on one math student-teacher pair and one code pair. Full training dynamics are provided in Appendix~\ref{app:training_dynamics}.}
\label{fig:training-dynamics}
\end{figure}

We present the main benchmark results in Table~\ref{tab:math-results} (math) and Table~\ref{tab:code-results} (code), and the training dynamics in Figure~\ref{fig:training-dynamics}.

\paragraph{Preserving the update direction is sufficient.}
Across all student-teacher pairs, BinaryOPD, which replaces the reverse KL objective with a simple binary directional reward, achieves performance comparable to or slightly better than OPD. For example, on the Qwen3-4B-Non-Thinking math pair, BinaryOPD reaches an average score of 65.1 versus 64.1 for OPD; on the Qwen3-1.7B-Base math pair, BinaryOPD achieves 22.2 versus 21.2. On the code benchmark, BinaryOPD and OPD are nearly identical (55.2 vs. 55.4). The training dynamics in Figure~\ref{fig:training-dynamics} further confirm this: BinaryOPD and OPD exhibit the same trend in training accuracy, entropy, and response length. This indicates that the directional signal alone---whether the teacher prefers a token more or less than the student---is sufficient to reproduce the OPD training mode, and that the magnitude information in reverse KL may be unnecessary.

\subsection{Why can BinaryOPD work? A repeated-feedback view.}
Unlike conventional distillation, OPD repeatedly obtains fresh teacher
supervision on states visited by the evolving student. Recent work has shown
that even a small number of queries can cover much of the state space visited
by full-data OPD, suggesting substantial redundancy in OPD
states~\citep{fu2026rethinking}. We further show in
Appendix~\ref{sec:state_saturation} that this redundancy is intrinsic to the
training process itself: state exploration saturates almost immediately, and
states encountered later in training repeatedly return to regions already
visited much earlier.

This changes how an OPD update should be interpreted. If a state were observed
only once, the magnitude of the teacher--student discrepancy would be needed
to determine how far the student should move. Under repeated visitation,
however, supervision becomes a closed-loop feedback process. OPD only
needs to indicate whether the student probability is above or below the
teacher probability. The same state region is then revisited after the student
has changed, producing a fresh directional signal; once the student crosses
the teacher, the direction automatically reverses. Repeated directional
feedback can therefore determine the required amount of correction
iteratively, without explicitly encoding its magnitude in any individual
update. From this view, the redundancy of OPD states makes the precise KL
magnitude not merely replaceable, but unnecessary.

\section{Only the Direction of High-Disagreement Tokens Is Critical for OPD}
\label{sec:high-disagreement}

Our previous results show that the update direction toward the teacher matters more than its magnitude. In this section, we show that only the direction of a small subset of tokens is critical: OPD works as long as the update direction of a very small subset of tokens with large teacher-student disagreement is toward the teacher.

\subsection{Method}
\label{sec:high-disagreement-method}

We use the log-ratio to represent the token-level disagreement between the teacher and the student: $\ell_t = \log \frac{\pi_T(o_t \mid q, o_{<t})}{\pi_\theta(o_t \mid q, o_{<t})}$.
A large positive \(\ell_t\) means the teacher strongly prefers this token over the student, while a large negative \(\ell_t\) means the student strongly overestimates it relative to the teacher. Based on a threshold \(\epsilon > 0\), we partition tokens into three groups:
\begin{itemize}
    \item \textbf{Group A} (high disagreement, teacher prefers more): \(\ell_t > \epsilon\).
    \item \textbf{Group B} (low disagreement): \(-\epsilon \le \ell_t \le \epsilon\).
    \item \textbf{Group C} (high disagreement, student prefers more): \(\ell_t < -\epsilon\).
\end{itemize}
Groups A and C together form the high-disagreement tokens, while Group B contains the vast majority of tokens with low disagreement.

To test which tokens are critical, we run two symmetric experiments with \(\epsilon = 0.8\). In the \textbf{positive reward} experiment, we assign \(+1\) uniformly to the selected tokens and consider three subsets: (i) only Group A, (ii) Groups A and B, (iii) Groups A, B, and C. In the \textbf{negative reward} experiment, we assign \(-1\) uniformly to the selected tokens and consider three subsets: (i) only Group C, (ii) Groups C and B, (iii) Groups C, B, and A. All unselected tokens receive \(0\). 

In the positive reward experiment, Group A tokens are those where the teacher probability is higher than the student probability, so assigning \(+1\) moves the student toward the teacher; Group C tokens are those where the student probability is higher than the teacher probability, so assigning \(+1\) moves the student away from the teacher. In the negative reward experiment, the roles are reversed.

\subsection{Results}
\label{sec:high-disagreement-results}
\begin{figure}[t]
    \centering
    \includegraphics[width=0.9\textwidth]{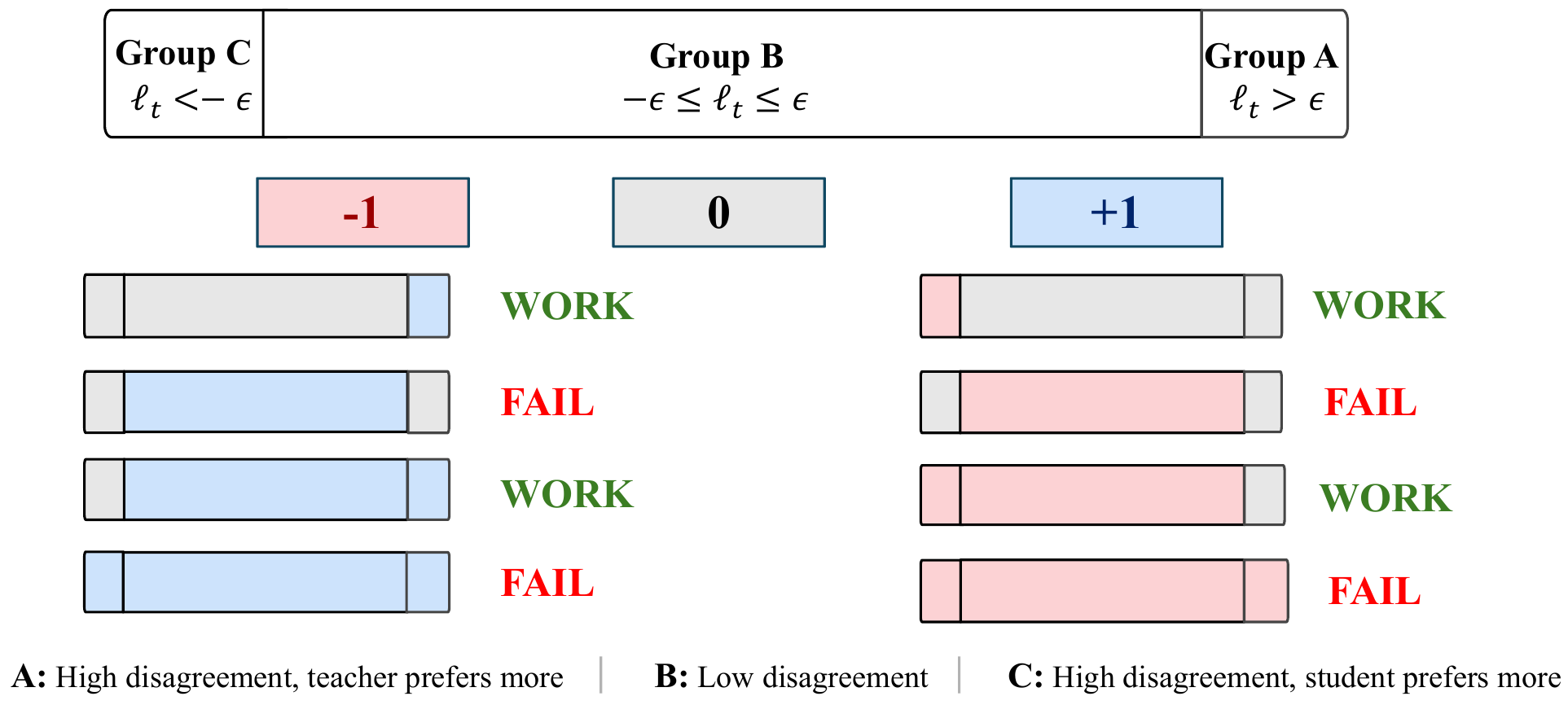}
    \caption{Summary of the positive and negative reward experiments. The positive experiment shows that keeping only Group A or Groups A+B leads to successful training, while adding Group C causes failure. The negative experiment exhibits a symmetric pattern.}
    \label{fig:positive-negative-summary}
\end{figure}
\begin{figure}[t]
    \centering
    \begin{minipage}{0.49\textwidth}
        \centering
        \includegraphics[width=\linewidth]{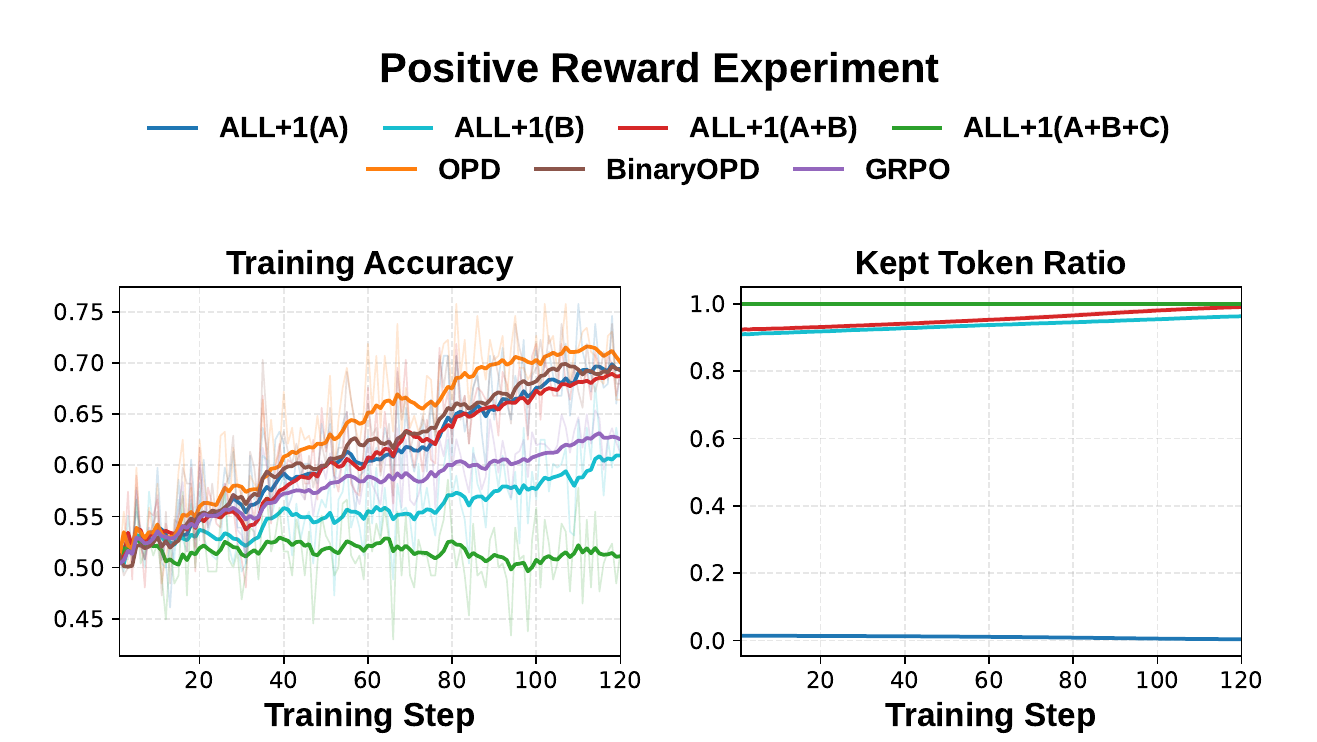}
        \centerline{(a) Positive reward experiment}
    \end{minipage}
    \hfill
    \begin{minipage}{0.49\textwidth}
        \centering
        \includegraphics[width=\linewidth]{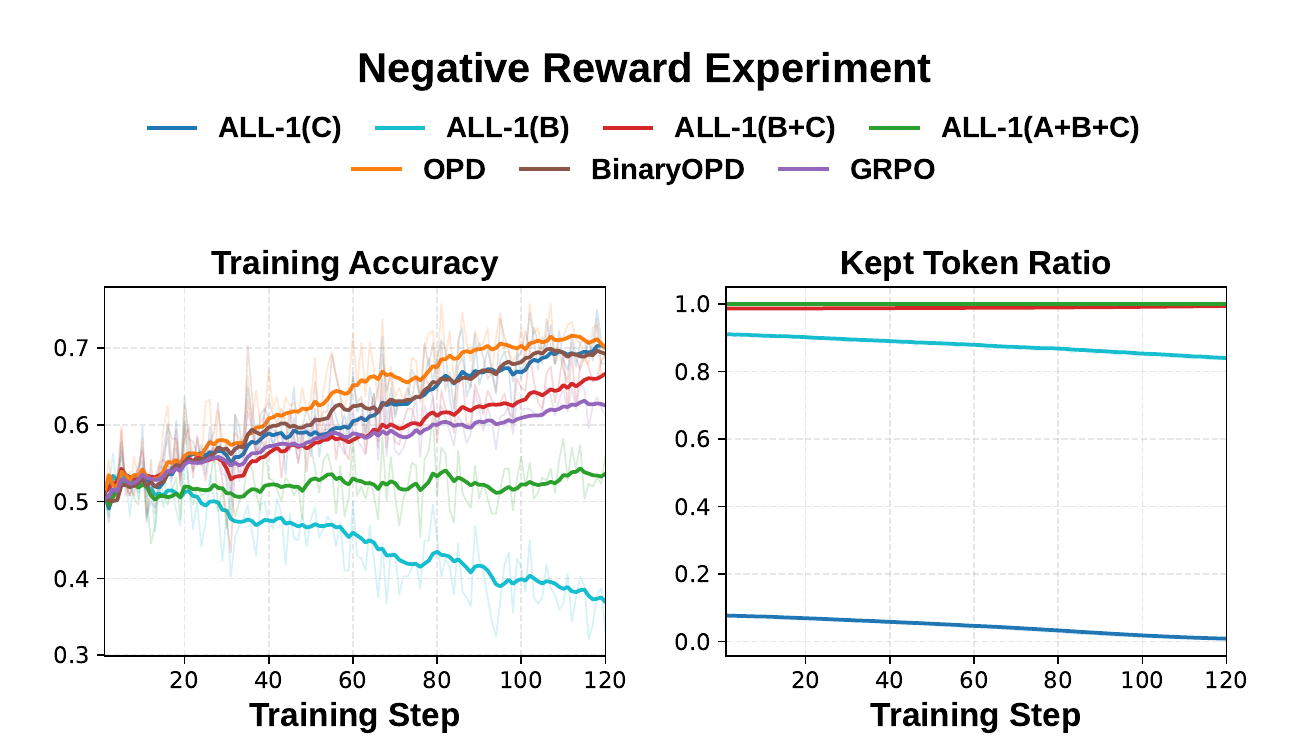}
        \centerline{(b) Negative reward experiment}
    \end{minipage}
    \caption{Training dynamics for the positive and negative reward experiments with \(\epsilon=0.8\), using Qwen3-4B-Non-Thinking-RL-Math as the teacher and Qwen3-4B-Non-Thinking as the student.}
    \label{fig:all-reward-experiments}
\end{figure}

\begin{table}[t]
\centering
\caption{Math benchmark results for the positive reward experiment (avg@8, \%). Note that \textbf{ALL+1(A) keeps less than \(1.5\%\) of all tokens.}}
\label{tab:positive-reward-results}
\small
\setlength{\tabcolsep}{4pt}
\renewcommand{\arraystretch}{1.15}
\begin{tabular}{@{}lccccccc@{}}
\toprule
Method & AIME24 & AIME25 & AMC & MATH500 & Minerva & Olympiad & Avg \\
\midrule
OPD & 60.4 & 51.7 & 85.7 & 93.0 & 34.5 & 59.2 & 64.1 \\
BinaryOPD & 61.3 & 55.0 & 85.7 & 92.8 & 35.9 & 59.8 & 65.1 \\
ALL+1(A) & 60.8 & 52.0 & 85.9 & 92.9 & 35.0 & 59.5 & 64.5 \\
ALL+1(A+B) & 60.5 & 51.5 & 86.0 & 92.7 & 34.8 & 59.0 & 64.3 \\
ALL+1(A+B+C) & 24.0 & 15.5 & 56.0 & 79.5 & 26.0 & 41.0 & 40.3 \\
\bottomrule
\end{tabular}
\end{table}

Figure~\ref{fig:all-reward-experiments} shows the training dynamics for the Qwen3-4B-Non-Thinking-RL-Math teacher and Qwen3-4B-Non-Thinking student with \(\epsilon = 0.8\). As shown, Group B (low disagreement) accounts for over \(90\%\) of all tokens, while Groups A and C (high disagreement) each account for only a small fraction. 

The results are surprising. In the positive reward experiment:
\begin{itemize}
    \item \textbf{Only Group A.} Training works. This setting keeps less than \(1.5\%\) of tokens (Group A) and assigns \(+1\) to them.
    \item \textbf{Only Group B.} Training improves but underperforms GRPO, which is also considered a failure for OPD's expected effectiveness.
    \item \textbf{Groups A and B.} Training works. This setting assigns \(+1\) to over \(90\%\) of the tokens (Groups A and B).
    \item \textbf{Groups A, B, and C.} Training fails. Compared with the working setting of Groups A and B, this setting only adds Group C, which accounts for less than \(5\%\) of all tokens.
\end{itemize}

In the negative reward experiment, the symmetric pattern holds:
\begin{itemize}
    \item \textbf{Only Group C.} Training works. This setting keeps only Group C and assigns \(-1\) to them.
     \item \textbf{Only Group B.} Training fails.
    \item \textbf{Groups C and B.} Training works and surpasses GRPO, although slightly worse than keeping only Group C. This setting assigns \(-1\) to over \(90\%\) of the tokens (Groups C and B).
    \item \textbf{Groups C, B, and A.} Training fails. Compared with the working setting of Groups C and B, this setting only adds Group A, which accounts for less than \(5\%\) of all tokens.
\end{itemize}

These results show that the critical tokens are the small fraction of high-disagreement tokens in Groups A and C. As long as their update direction is toward the teacher, OPD works. Once their direction is pulled away from the teacher, OPD fails. In contrast, the vast majority of low-disagreement tokens in Group B are not critical: removing them does not break OPD, and including them with an incorrect reward direction may cause some degradation but still allows training to proceed, as long as the high-disagreement tokens are handled correctly. We also conduct experiments with \(\epsilon = 0.2\) and \(0.5\), as well as another student-teacher pair (JustRL-1.5B and DeepSeek-Distill-Qwen-1.5B). Complete experiments are provided in Appendix~\ref{app:training_dynamics}.

\section{Consensus Multi-Teacher On-Policy Distillation}
\label{sec:c-mopd}

\subsection{Method}
\label{sec:c-mopd-method}

Our observation that OPD only needs updates toward the teacher, and that the majority of tokens, which do not have large teacher-student disagreement, can be pulled away from the teacher without hurting performance, suggests that in the multi-teacher setting we need not route each sample to a single teacher. Instead, we can guide every token with all teachers simultaneously, and when teachers conflict, we can still update toward the domain-specific teacher while ensuring that this update does not hurt the capabilities associated with other teachers' directions. Based on this idea, we propose \textbf{Consensus Multi-Teacher On-Policy Distillation (C-MOPD)}.

Let \(\{\pi_{T_1}, \dots, \pi_{T_K}\}\) be \(K\) teachers. For a query \(q\), the student generates a rollout \(o \sim \pi_\theta(\cdot \mid q)\). At each token \(o_t\), let
\[
p_s = \pi_\theta(o_t \mid q, o_{<t}), \qquad
p_k = \pi_{T_k}(o_t \mid q, o_{<t}), \quad k = 1, \dots, K.
\]
Let \(k^*\) be the index of the domain-specific teacher for this sample, and define
\[
\ell_k = \log(p_k / p_s), \qquad d = \operatorname{sign}(\ell_{k^*}) \in \{+1, -1\}.
\]
We assign the reward as follows:
\[
r_t =
\begin{cases}
+1, & \text{if } p_k > p_s \ \forall k, \\[2pt]
-1, & \text{if } p_k < p_s \ \forall k, \\[2pt]
+1, & \text{if teachers conflict, } d = +1, \text{ and } \ell_k > -\epsilon \ \forall k \neq k^*, \\[2pt]
-1, & \text{if teachers conflict, } d = -1, \text{ and } \ell_k < \epsilon \ \forall k \neq k^*, \\[2pt]
0,  & \text{otherwise}.
\end{cases}
\]

This design ensures that each update either pulls toward all teachers, or toward the domain-specific teacher without hurting the capabilities associated with the other teachers' directions.

\subsection{Experiments}
\label{sec:c-mopd-experiments}

\subsubsection{Experimental Setup}
\label{sec:c-mopd-setup}

We evaluate C-MOPD in a multi-teacher setting with two domain-specific teachers: a math teacher and a code teacher. The student model is Qwen3-4B-Non-Thinking. The math teacher is Qwen3-4B-Non-Thinking-RL-Math, and the code teacher is Qwen3-4B-Non-Thinking-RL-Code. The training set is a mixture of math and code problems, containing 25,276 DeepMath problems, 9,639 CodeContests problems, 9,579 TACO problems, 3,462 APPS problems, and 2,596 Codeforces problems. We compare C-MOPD against MOPD. Based on our empirical findings in Section~\ref{sec:high-disagreement}, we set \(\epsilon=0.8\) in C-MOPD by default. We also compare with a conservative setting of \(\epsilon=0.0\).

\subsubsection{Experimental Results}

\begin{table}[t]
\centering
\caption{Math and code benchmark results for C-MOPD (avg@8, \%).}
\label{tab:c-mopd-all}
\footnotesize
\setlength{\tabcolsep}{3pt}
\renewcommand{\arraystretch}{1.15}
\resizebox{\linewidth}{!}{%
\begin{tabular}{@{}l*{9}{c}@{}}
\toprule
& \multicolumn{6}{c}{\textbf{Math}} & \multicolumn{3}{c}{\textbf{Code}} \\
\cmidrule(lr){2-7}\cmidrule(l){8-10}
\textbf{Model}
& \textbf{AIME24} & \textbf{AIME25} & \textbf{AMC}
& \textbf{MATH500} & \textbf{Minerva} & \textbf{Olympiad}
& \textbf{LCB v6} & \textbf{HumanEval} & \textbf{MBPP} \\
\midrule
\rowcolor{baselinegray}
Student & 25.0 & 16.7 & 58.4 & 81.3 & 27.7 & 42.9 & 25.1 & 80.8 & 48.7 \\
\rowcolor{baselinegray}
Math Teacher & 61.2 & 56.3 & 88.0 & 93.2 & 36.7 & 60.5 & -- & -- & -- \\
\rowcolor{baselinegray}
Code Teacher & -- & -- & -- & -- & -- & -- & 36.0 & 90.6 & 56.2 \\
\addlinespace[2pt]
\rowcolor{cplblue}
MOPD & 60.0 & 52.5 & 86.4 & 92.9 & 34.3 & 58.8 & 27.9 & 85.9 & 53.5 \\
\rowcolor{cplblue}
\textbf{C-MOPD (\(\epsilon=0.0\))} & 59.8 & 52.8 & 86.2 & 92.8 & 34.6 & 58.9 & 28.2 & 85.7 & 53.8  \\
\rowcolor{cplblue}
\textbf{C-MOPD (\(\epsilon=0.8\))} 
& \textbf{61.3} & \textbf{53.9} & \textbf{88.0}
& \textbf{93.0} & \textbf{35.5} & \textbf{60.4}
& \textbf{32.9} & \textbf{87.8} & \textbf{54.3} \\
\bottomrule
\end{tabular}
}
\end{table}
\begin{figure}[h]
    \centering
    \includegraphics[width=\textwidth]{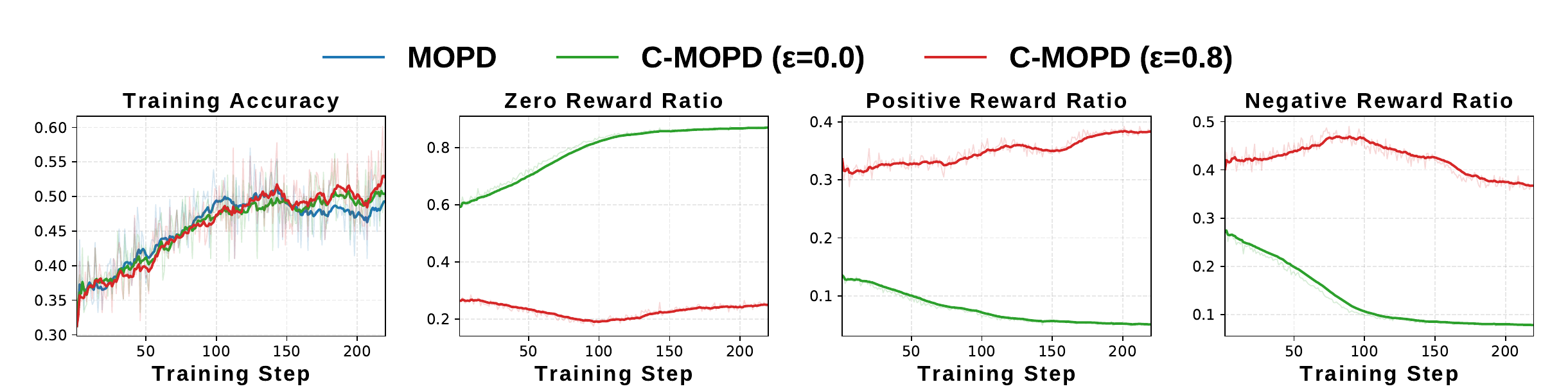}
    \caption{Training dynamics of C-MOPD.}
    \label{fig:mopd-vs-cmopd}
\end{figure}

Table~\ref{tab:c-mopd-all} shows that C-MOPD with \(\epsilon=0.8\) consistently outperforms MOPD across almost all math and code benchmarks. In contrast, \(\epsilon=0.0\) is overly conservative: its Zero Reward Ratio increases monotonically and eventually exceeds \(90\%\), meaning that as training proceeds, an increasing number of tokens become conflicting across teachers, so the model refuses to update on most of them, and its performance only matches MOPD. This result reinforces our earlier finding: tokens with small teacher-student disagreement can be updated away from the teacher without severely hurting performance. By allowing updates toward the domain teacher when the conflict is mild, \(\epsilon=0.8\) exploits this observation and achieves the best overall results.

\section{Conclusion}
\label{sec:conclusion}

In this work, we revisit the role of reverse KL divergence in on-policy distillation (OPD). We show that KL divergence may not be necessary: a simple binary directional reward, which only ensures updates toward the teacher, reproduces OPD-like training and achieves comparable or better results. We further find that only a small subset of tokens with large teacher-student disagreement is critical, and that the verifier signal has little effect on OPD. As an application, we propose C-MOPD, which consistently outperforms MOPD on both math and code benchmarks. We hope this work encourages a rethinking of KL's role in LLM post-training and opens the door to simpler distillation objectives.

\FloatBarrier

\newpage
\bibliography{main}
\bibliographystyle{icml2025}

\appendix
\section{State Exploration Is Highly Redundant in OPD Training}
\label{sec:state_saturation}
\begin{figure}[t]
    \centering
    \includegraphics[width=0.98\linewidth]{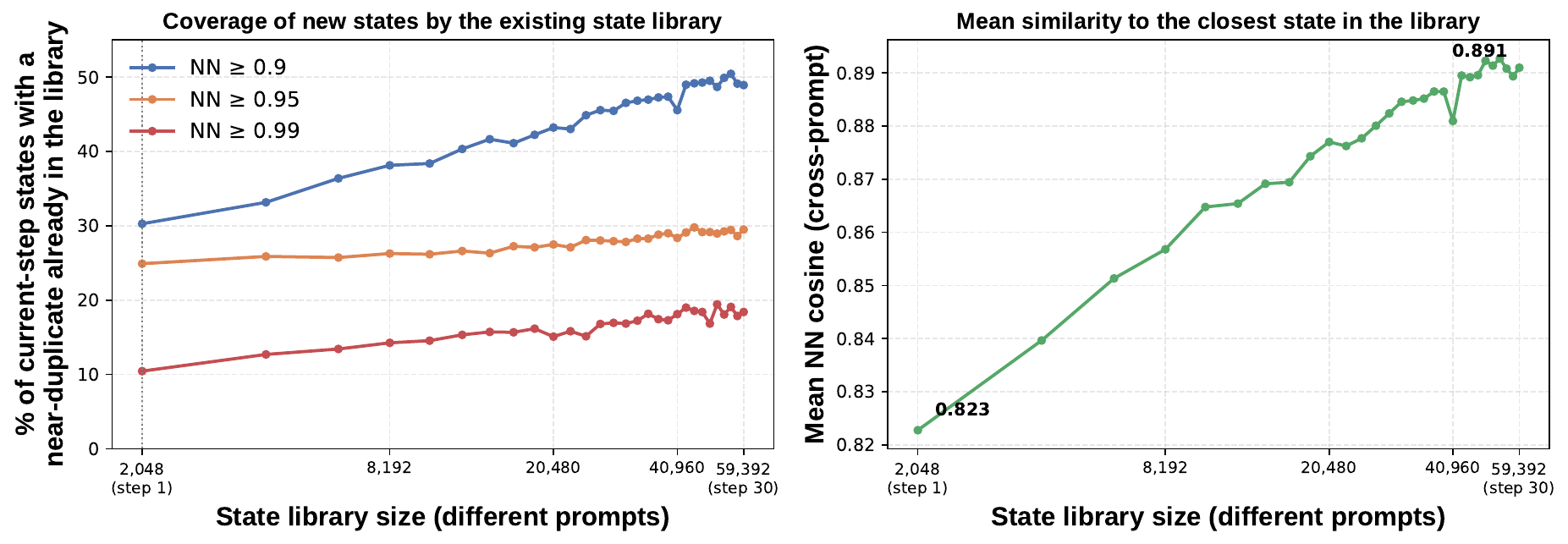}
    \caption{\textbf{State-space coverage grows slowly despite a rapidly
    expanding state library.}
    For each current state, we search only among previously visited states from
    different prompts. \emph{Left:} fraction of current-step states whose
    closest earlier neighbor exceeds cosine thresholds $0.9$, $0.95$, and
    $0.99$, as a function of library size. Expanding the library by $29\times$
    raises $\cos\geq0.99$ coverage from $10.4\%$ to $18.4\%$.
    \emph{Right:} mean similarity to the closest earlier state increases from
    $0.823$ to $0.891$.}
    \label{fig:saturation}
\end{figure}

\begin{figure}[t]
    \centering
    \includegraphics[width=0.9\linewidth]{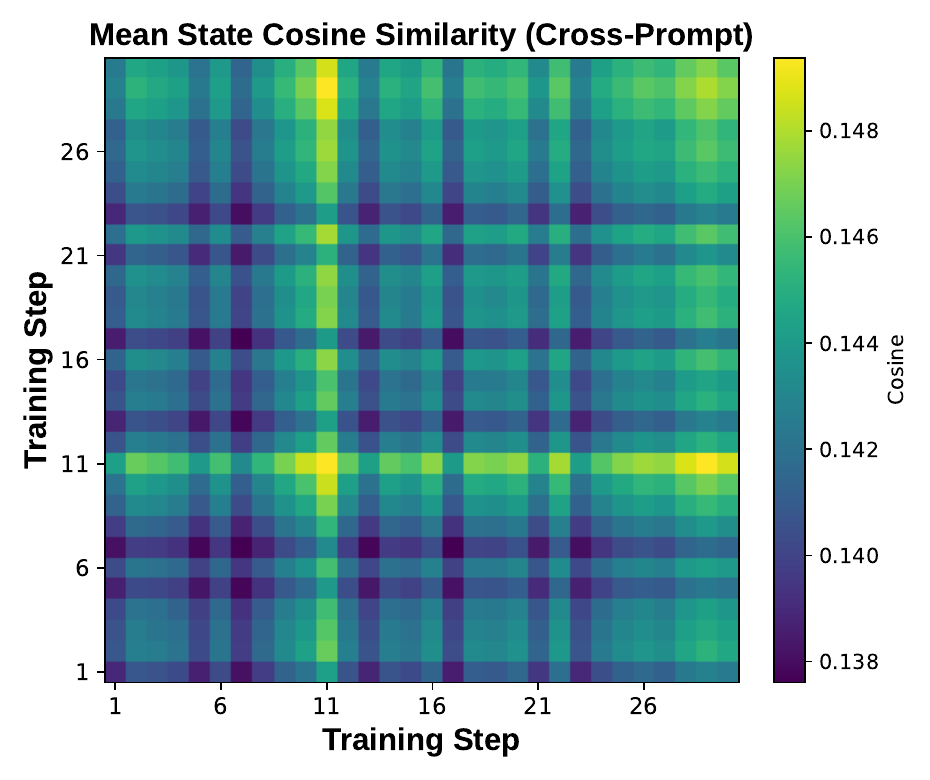}
    \caption{\textbf{The geometry of visited states remains stable across
    training.}
    Mean cross-prompt cosine similarity between states collected at training
    steps $i$ and $j$. All off-diagonal entries lie in the narrow range
    $[0.138, 0.149]$, with no clear separation between early and late training
    steps.}
    \label{fig:step_heatmap}
\end{figure}

\begin{figure}[t]
    \centering
    \includegraphics[width=0.98\linewidth]{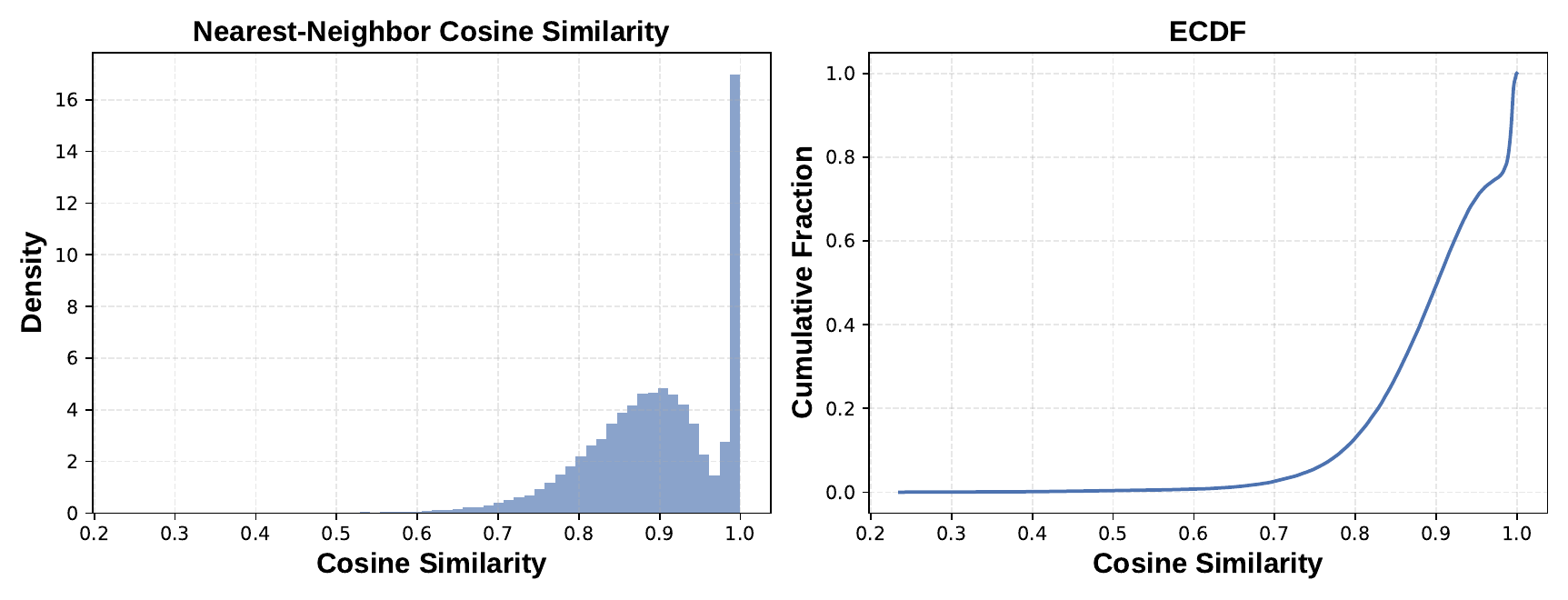}
    \caption{\textbf{Strong cross-prompt nearest-neighbor redundancy.}
    Distribution of nearest-neighbor cosine similarity over all $61{,}440$
    states, where the neighbor is required to originate from a different
    prompt. The mean nearest-neighbor similarity is $0.893$; $29.5\%$ of
    states have a neighbor with cosine similarity at least $0.95$, and
    $18.5\%$ have one at least $0.99$.}
    \label{fig:nn_dist}
\end{figure}
We show that OPD training visits a highly redundant state space that changes little over training. Each state is unique at the token level, but states from different training steps and prompts often occupy nearby regions in a fixed teacher representation space. Increasing the number of observed states gives only small gains in covering future states. This suggests limited expansion of the represented state space during training.

\paragraph{Setup.}
We instrument OPD training to record the states visited by on-policy rollouts. We use a Qwen3-4B student distilled from Qwen3-4B-Non-Thinking-RL-Math on DeepMath. A state is the full autoregressive context: the prompt \(x\) and the student-generated prefix \(y_{<t}\), denoted \(s_t = (x, y_{<t})\). Over 30 training steps we collect 61,440 states from 7,680 distinct prompts, with 256 rollouts per step and 8 uniformly spaced positions per rollout. Each state is unique at the token level. To compare states across training, we map each state into a fixed representation space using the frozen teacher: \(h_T(s)\) is the teacher's final-layer hidden state at the last token of \(s\). All similarities are cosine similarities in this space. Since the teacher is fixed, representations from different steps and prompts are directly comparable.

\paragraph{Strong cross-prompt redundancy.}
For each state, we find its nearest neighbor among states from a different prompt (Figure~\ref{fig:nn_dist}). Even with this restriction, the nearest-neighbor similarity is high. The mean is \(0.893\) and the median is \(0.901\). The upper tail is concentrated: the 90th and 95th percentiles reach \(0.994\) and \(0.995\). Overall, \(29.5\%\) of states have a different-prompt neighbor with cosine similarity at least \(0.95\), and \(18.5\%\) have one at least \(0.99\) (Table~\ref{tab:nn_stats}). Restricting to different prompts leaves the mean nearest-neighbor similarity almost unchanged: \(0.893\) versus \(0.894\) without this restriction. Thus, this redundancy is not just from repeated prefixes of the same prompt. States from different training problems often lie in nearby regions of the teacher representation space.

\begin{table}[t]
\centering
\small
\caption{Cross-prompt nearest-neighbor similarity over all $61{,}440$ states.
The last two columns report the fraction of states whose nearest
\emph{different-prompt} neighbor exceeds the corresponding threshold.}
\label{tab:nn_stats}
\begin{tabular}{lcccccc}
\toprule
 & mean & median & p90 & p95 & $\%\geq 0.99$ & $\%\geq 0.95$ \\
\midrule
Cross-prompt NN & 0.893 & 0.901 & 0.994 & 0.995 & 18.5\% & 29.5\% \\
\bottomrule
\end{tabular}
\end{table}

\paragraph{The visited-state geometry remains stable across training.}
We next ask whether later training steps occupy systematically different regions of the representation space. Figure~\ref{fig:step_heatmap} shows the mean cross-prompt cosine similarity between states from every pair of training steps. The matrix is nearly flat: all off-diagonal entries fall within a narrow range \([0.138, 0.149]\). In particular, the similarity between step 1 and step 30 is comparable to that between nearby steps. A nearest-neighbor analysis agrees: the fraction of states whose nearest neighbor comes from the same training step is about the chance level of \(1/30\). Later states do not tend to cluster with other late-training states rather than with much earlier states. These results do not mean the exact state distribution is unchanged. Rather, under this fixed representation, training does not produce a clear progressive separation of visited states into new regions.

\paragraph{A small early state library already covers a substantial fraction of later states.}
Figure~\ref{fig:saturation} quantifies this redundancy. For each state at a given step, we search only among states from earlier steps and different prompts. The 2,048 states from step 1 already give a neighbor with cosine similarity at least 0.99 for 10.4\% of states in the next 29 steps, and at least 0.95 for 24.9\%. Expanding the library by 29\(\times\), from 2,048 to 59,392 states, adds only modest coverage: the fraction with a \(\cos \ge 0.99\) neighbor rises from 10.4\% to 18.4\%, and the mean similarity to the closest earlier state goes from 0.823 to 0.891. So even though the library grows a lot, the representational novelty from more training steps grows much more slowly.

Taken together, these results show substantial redundancy in the states visited during OPD. Training keeps generating token-level distinct contexts, but many of them are close to states already seen under other prompts. This repeated coverage helps explain why OPD can stay effective even when each update keeps only coarse directional information instead of the precise magnitude of the teacher-student discrepancy. Because the same or nearby states are visited repeatedly, the student is corrected many times in the same direction, and BinaryOPD can work despite discarding all magnitude information.

\section{Whether the OPD Update Direction Is Aligned with the Verifier Is Not Important}\label{app:verifier}
Recently, combining OPD with RLVR has become a popular direction. OPD provides dense teacher guidance but does not consider trajectory correctness, while RLVR~\citep{DeepSeekR1,DeepSeekMath,jaech2024openai,trinh2024solving,yang2024qwen2} optimizes directly toward trajectory-level correctness but suffers from sparse rewards. Combining them therefore appears to be an effective and promising direction~\citep{wang2026distilled,cai2026h,hou2026uni,yang2026self,akhondzadeh2026reward,lin2026policy,ding2026saf}. As a side investigation, we explore whether the verifier signal is important for OPD. Our setup builds directly on BinaryOPD. In the standard RLVR setting, trajectories with correct final answers are positively reinforced, while trajectories with incorrect final answers are negatively reinforced. Here, we completely reverse this relationship. Specifically:

\begin{itemize}
    \item On trajectories with \textbf{correct} final answers, we keep only tokens whose teacher probability is \emph{lower} than the student probability, and assign them a reward of \(-1\).
    \item On trajectories with \textbf{incorrect} final answers, we keep only tokens whose teacher probability is \emph{higher} than the student probability, and assign them a reward of \(+1\).
\end{itemize}

As a result, tokens in correct trajectories receive non-positive rewards, while tokens in incorrect trajectories receive non-negative rewards. This is the exact opposite of what a verifier would normally encourage.

Formally, for a rollout \(o\) with final-answer correctness \(c \in \{0,1\}\), the reward at token \(o_t\) is
\[
r_t =
\begin{cases}
-1, & \text{if } c = 1 \text{ and } \pi_T(o_t \mid q, o_{<t}) < \pi_\theta(o_t \mid q, o_{<t}), \\[2pt]
+1, & \text{if } c = 0 \text{ and } \pi_T(o_t \mid q, o_{<t}) > \pi_\theta(o_t \mid q, o_{<t}), \\[2pt]
0,  & \text{otherwise}.
\end{cases}
\]
We denote the reversed-mask variant as \textbf{Verifier-Conflicted}. For comparison, we also evaluate \textbf{Verifier-Aligned}, which keeps only tokens whose update direction aligns with the verifier signal.

If the verifier signal were beneficial for OPD, this reversed masking should severely degrade performance or completely change the training dynamics, and Verifier-Aligned should yield improvements over Verifier-Conflicted and OPD.

\begin{table}[h]
\centering
\caption{Math benchmark results. All results are averaged over 8 samples (Avg@8). }
\label{app:math-results}
\small
\setlength{\tabcolsep}{4pt}
\begin{tabular}{@{}l*{7}{c}@{}}
\toprule
\textbf{Method} & \textbf{AIME24} & \textbf{AIME25} & \textbf{AMC}
& \textbf{MATH500} & \textbf{Minerva} & \textbf{Olympiad} & \textbf{Avg} \\
\midrule
\multicolumn{8}{@{}l}{\textit{DeepSeek-Distill-Qwen-1.5B (Student) vs. JustRL-1.5B (Teacher)}} \\
\addlinespace[2pt]
\rowcolor{baselinegray}
Student & 23.7 & 17.0 & 59.7 & 80.9 & 23.7 & 41.4 & 41.0 \\
\rowcolor{baselinegray}
Teacher & 46.2 & 37.5 & 82.0 & 89.2 & 30.7 & 53.5 & 56.5 \\
\addlinespace[2pt]
\rowcolor{cplblue}
OPD & 43.8 & 32.5 & 78.8 & 88.5 & 31.7 & 52.6 & 54.7 \\
\rowcolor{cplblue}
BinaryOPD & \textbf{45.4} & 32.1 & 78.5 & \textbf{88.9} & 31.6 & 52.6 & 54.9 \\
\rowcolor{cplblue}
Verifier-Aligned & 44.6 & 31.3 & 78.0 & 88.7 & 31.3 & 52.6 & 54.4 \\
\rowcolor{cplblue}
Verifier-Conflicted & 45.0 & \textbf{32.9} & \textbf{79.2} & 88.8 & \textbf{31.8} & \textbf{53.0} & \textbf{55.1} \\
\midrule
\multicolumn{8}{@{}l}{\textit{Qwen3-1.7B-Base (Student) vs. Qwen3-4B-Base-RL (Teacher)}} \\
\addlinespace[2pt]
\rowcolor{baselinegray}
Student & 4.1 & 1.7 & 23.2 & 48.9 & 8.9 & 17.1 & 17.3 \\
\rowcolor{baselinegray}
Teacher & 10.6 & 13.1 & 40.3 & 74.2 & 17.2 & 30.0 & 30.9 \\
\addlinespace[2pt]
\rowcolor{cplblue}
OPD & 7.5 & 3.3 & 26.4 & 57.8 & 11.4 & 20.6 & 21.2 \\
\rowcolor{cplblue}
BinaryOPD & \textbf{8.1} & \textbf{4.0} & 30.3 & \textbf{59.1} & 10.0 & 21.6 & \textbf{22.2} \\
\rowcolor{cplblue}
Verifier-Aligned & 5.0 & 2.9 & 29.5 & 58.5 & 11.0 & 21.4 & 21.4 \\
\rowcolor{cplblue}
Verifier-Conflicted & 6.5 & 2.3 & \textbf{31.2} & 57.3 & 11.3 & \textbf{23.4} & 22.0 \\
\midrule
\multicolumn{8}{@{}l}{\textit{Llama-3.2-3B-Instruct (Student) vs. GT-Llama3.2-3B-MATH (Teacher)}} \\
\addlinespace[2pt]
\rowcolor{baselinegray}
Student & 2.9 & 0.0 & 14.8 & 34.5 & 7.7 & 7.9 & 11.3 \\
\rowcolor{baselinegray}
Teacher & 11.7 & 0.0 & 20.9 & 49.1 & 15.7 & 15.4 & 18.8 \\
\addlinespace[2pt]
\rowcolor{cplblue}
OPD & \textbf{12.5} & 0.4 & \textbf{22.7} & 46.0 & 13.3 & 14.3 & \textbf{18.2} \\
\rowcolor{cplblue}
BinaryOPD & 9.2 & 0.0 & 21.7 & 47.2 & 14.8 & 14.6 & 17.9 \\
\rowcolor{cplblue}
Verifier-Aligned & 8.3 & \textbf{0.8} & 20.8 & 46.3 & \textbf{15.3} & 14.2 & 17.6 \\
\rowcolor{cplblue}
Verifier-Conflicted & 9.2 & 0.0 & 21.7 & \textbf{47.4} & 15.2 & \textbf{14.8} & 18.1 \\
\midrule
\multicolumn{8}{@{}l}{\textit{Qwen3-4B-Non-Thinking (Student) vs. Qwen3-4B-Non-Thinking-RL-Math (Teacher)}} \\
\addlinespace[2pt]
\rowcolor{baselinegray}
Student & 25.0 & 16.7 & 58.4 & 81.3 & 27.7 & 42.9 & 42.0 \\
\rowcolor{baselinegray}
Teacher & 61.2 & 56.3 & 88.0 & 93.2 & 36.7 & 60.5 & 66.0 \\
\addlinespace[2pt]
\rowcolor{cplblue}
OPD & 60.4 & 51.7 & 85.7 & 93.0 & 34.5 & 59.2 & 64.1 \\
\rowcolor{cplblue}
BinaryOPD & \textbf{61.3} & \textbf{55.0} & \textbf{85.7} & 92.8 & \textbf{35.9} & \textbf{59.8} & \textbf{65.1} \\
\rowcolor{cplblue}
Verifier-Aligned & \textbf{61.3} & \textbf{55.0} & 84.6 & 92.7 & 34.6 & 59.3 & 64.6 \\
\rowcolor{cplblue}
Verifier-Conflicted & \textbf{61.3} & 51.7 & 84.9 & \textbf{93.2} & 34.4 & \textbf{59.8} & 64.2 \\
\bottomrule
\end{tabular}
\end{table}

\begin{table}[h]
\centering
\caption{Code benchmark results. All results are averaged over 8 samples (Avg@8).}
\label{app:code-results}
\small
\setlength{\tabcolsep}{4pt}
\begin{tabular}{@{}l*{4}{c}@{}}
\toprule
\textbf{Method} & \textbf{LiveCodeBench v6} & \textbf{HumanEval} & \textbf{MBPP} & \textbf{Avg} \\
\midrule
\multicolumn{5}{@{}l}{\textit{Qwen3-4B-Non-Thinking (Student) vs. Qwen3-4B-Non-Thinking-RL-Code (Teacher)}} \\
\addlinespace[2pt]
\rowcolor{baselinegray}
Student & 25.1 & 80.8 & 48.7 & 51.5 \\
\rowcolor{baselinegray}
Teacher & 36.0 & 90.6 & 56.2 & 60.9 \\
\addlinespace[2pt]
\rowcolor{cplblue}
OPD & 27.9 & 85.4 & 53.0 & \textbf{55.4} \\
\rowcolor{cplblue}
BinaryOPD & 27.8 & \textbf{85.7} & 52.1 & 55.2 \\
\rowcolor{cplblue}
Verifier-Aligned & 28.1 & 84.6 & \textbf{52.2} & 55.0 \\
\rowcolor{cplblue}
Verifier-Conflicted & \textbf{29.1} & 84.1 & 51.9 & 55.0 \\
\bottomrule
\end{tabular}
\end{table}

Tables~\ref{app:math-results} and~\ref{app:code-results} show that reversing the verifier signal does not hurt BinaryOPD. Across all student-teacher pairs, Verifier Conflicted performs on par with or even slightly better than OPD and BinaryOPD. For example, on the DeepSeek-Distill-Qwen-1.5B pair, Verifier Conflicted achieves the best average score of \(55.1\), while Verifier Aligned gets \(54.4\). On the Qwen3-4B-Non-Thinking math pair, all methods are close, with Verifier Conflicted and Verifier Aligned both reaching around \(64\). On code, Verifier Conflicted even achieves the highest LiveCodeBench v6 score (\(29.1\)). In contrast, Verifier Aligned shows no consistent improvement over OPD or BinaryOPD. As further shown in Figure~\ref{fig:app-training-dynamics} (Appendix~\ref{app:training_dynamics}), all methods exhibit nearly identical training dynamics, confirming that the verifier signal has little influence on the training pattern. Even if the BinaryOPD signal completely contradicts the verifier, the performance remains unaffected. These results clearly indicate that whether the update direction aligns with the verifier signal has little effect on BinaryOPD performance, questioning the necessity of incorporating the verifier signal into OPD.

\section{Experimental Details}\label{app:Experimental_details}
\paragraph{Models and Datasets.}
We use JustRL-1.5B following~\citep{he2025justrl}, Qwen3-4B-Non-Thinking-RL-Math and Qwen3-4B-Non-Thinking-RL-Code following~\citep{yang2026learning}, GT-Llama3.2-3B-MATH following~\citep{zhang2025coreward}, Nemotron-Research-Reasoning-1.5B following ~\citep{liu2026prorl} and Qwen3-4B-Base-RL. We obtain Qwen3-4B-Base-RL by training Qwen3-4B-Base with GRPO on DAPO-MATH-17K for 3 epochs. All these teachers are domain-specialized models obtained by running reinforcement learning on their respective domains. The DeepMath dataset follows~\citep{yang2026learning}, which selects 57K samples with a difficulty level greater than or equal to 6.
\paragraph{Hyperparameters.}
We use the \textbf{Verl} framework for training~\citep{verl}. We use the hyperparameters listed in Table~\ref{tab:hyperparameters} for all experiments. 
\begin{table}[h]
    \centering
    \caption{Hyperparameter settings.}
    \label{tab:hyperparameters}
    \vspace{5pt}
    \begin{tabular}{l|c}
        \toprule
        \textbf{Hyperparameter} & \textbf{Value} \\
        \midrule
        Learning Rate & 1e-6 \\
        Train Batch Size & 256 \\
        Max Response Length (Training) & 8192 \\
        Max Response Length (Evaluation) & 16384 \\
        Max Prompt Length & 1024 \\
        Rollout Temperature & 1.0 \\
        Evaluation Temperature & 0.7 \\
        Evaluation Top-$p$ & 0.95 \\
        \bottomrule
    \end{tabular}
\end{table}

\section{Training Dynamics}
\label{app:training_dynamics}

\begin{figure}[h]
    \centering
    \includegraphics[width=0.87\textwidth]{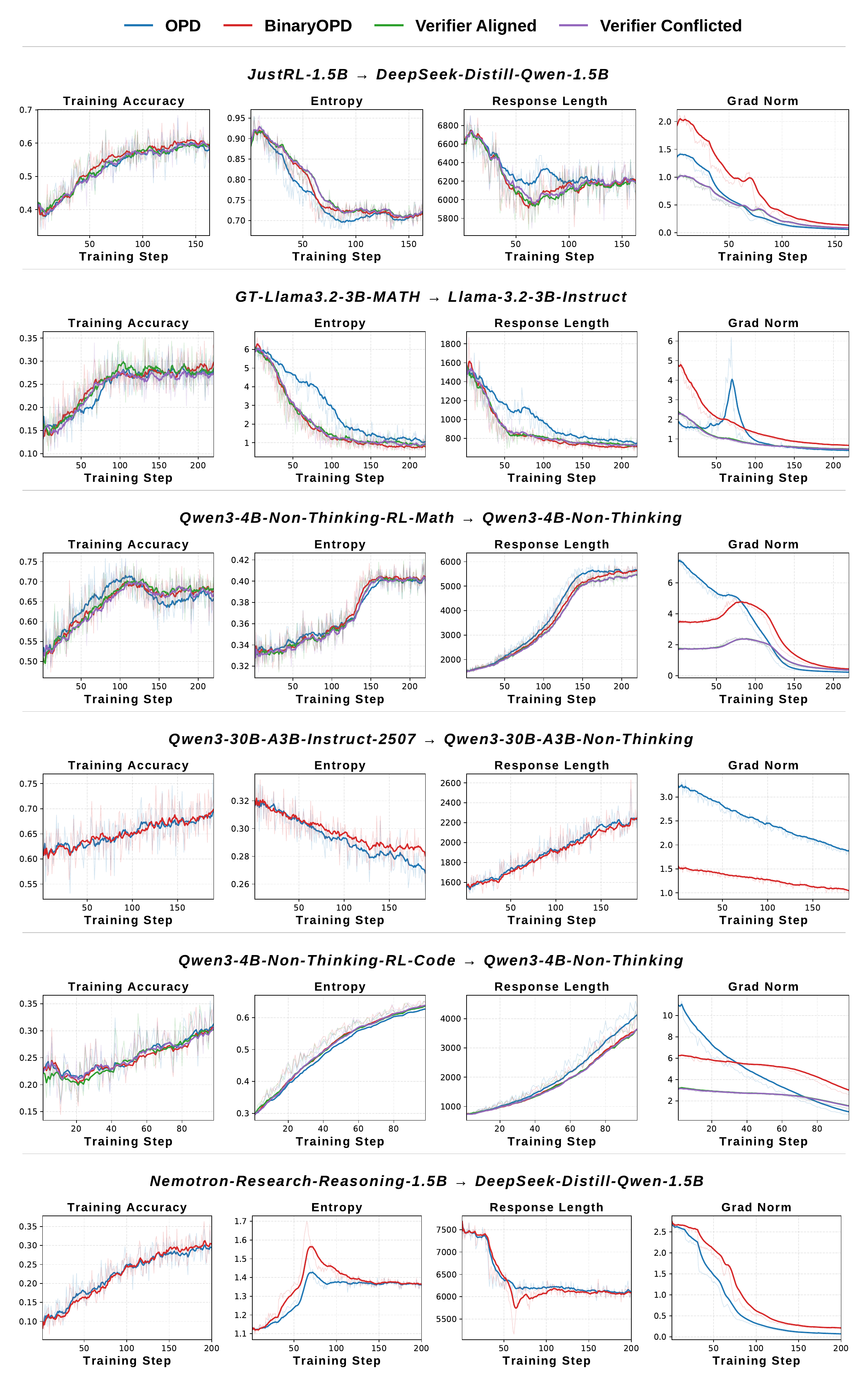}
    \caption{Full training dynamics of all methods (OPD, BinaryOPD, Verifier-Aligned, Verifier-Conflicted) across all student-teacher pairs and datasets. From left to right: training accuracy, entropy, response length, and gradient norm.}
    \label{fig:app-training-dynamics}
\end{figure}

Figure~\ref{fig:app-training-dynamics} shows the full training dynamics of all methods (OPD, BinaryOPD, Verifier Aligned, Verifier Conflicted) in the experiments of Section~\ref{sec:BinaryOPD} and Appendix~\ref{app:verifier}. Across all methods, the curves are nearly identical in training accuracy, entropy, response length, and gradient norm, confirming that removing the KL magnitude or reversing the verifier signal does not change the training pattern.

\begin{figure}[h]
    \centering
    \includegraphics[width=0.87\textwidth]{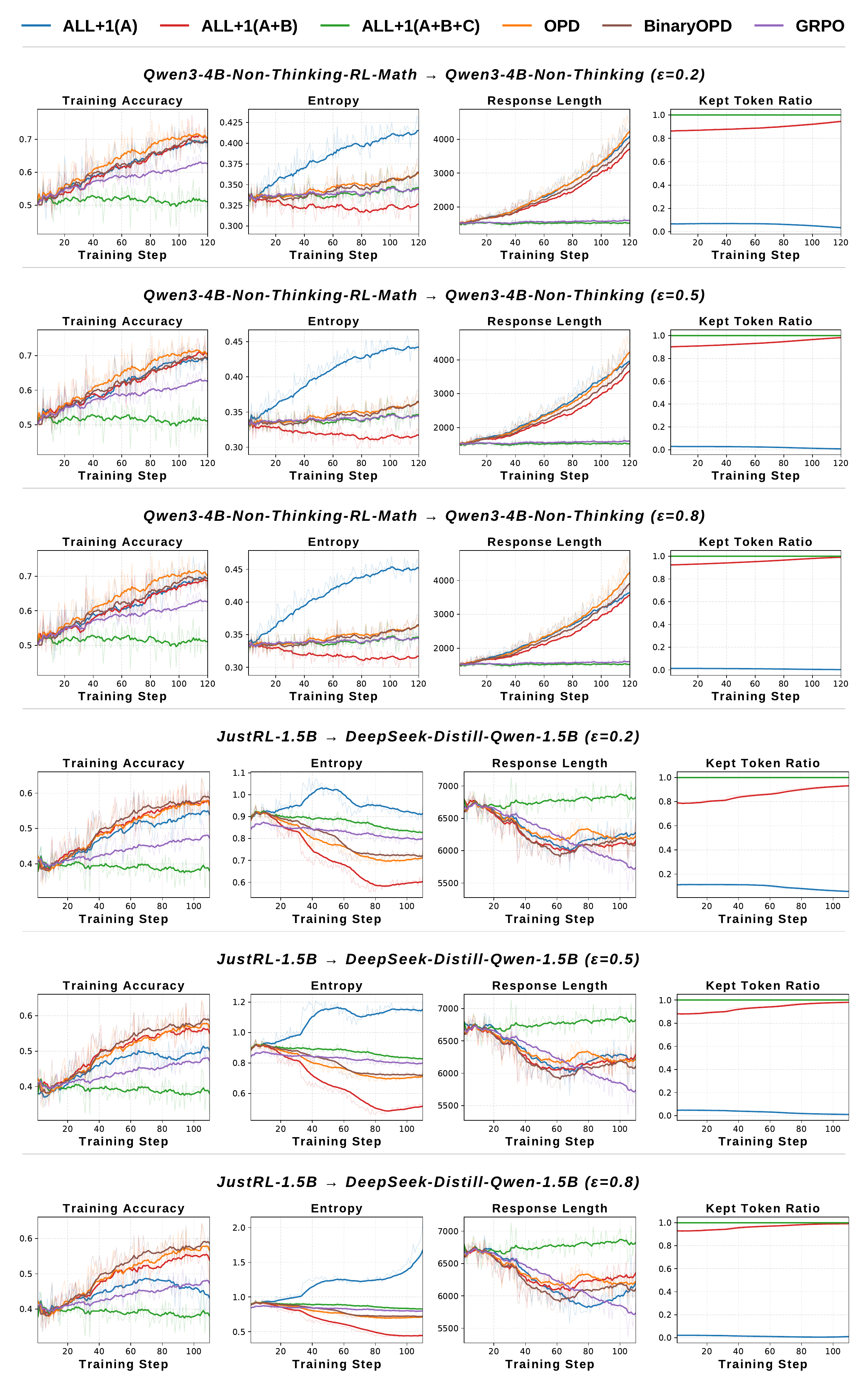}
    \caption{Full training dynamics for the positive reward experiment (ALL+1) across different \(\epsilon\) values and student-teacher pairs.}
    \label{fig:app-positive-dynamics}
\end{figure}

\begin{figure}[h]
    \centering
    \includegraphics[width=0.87\textwidth]{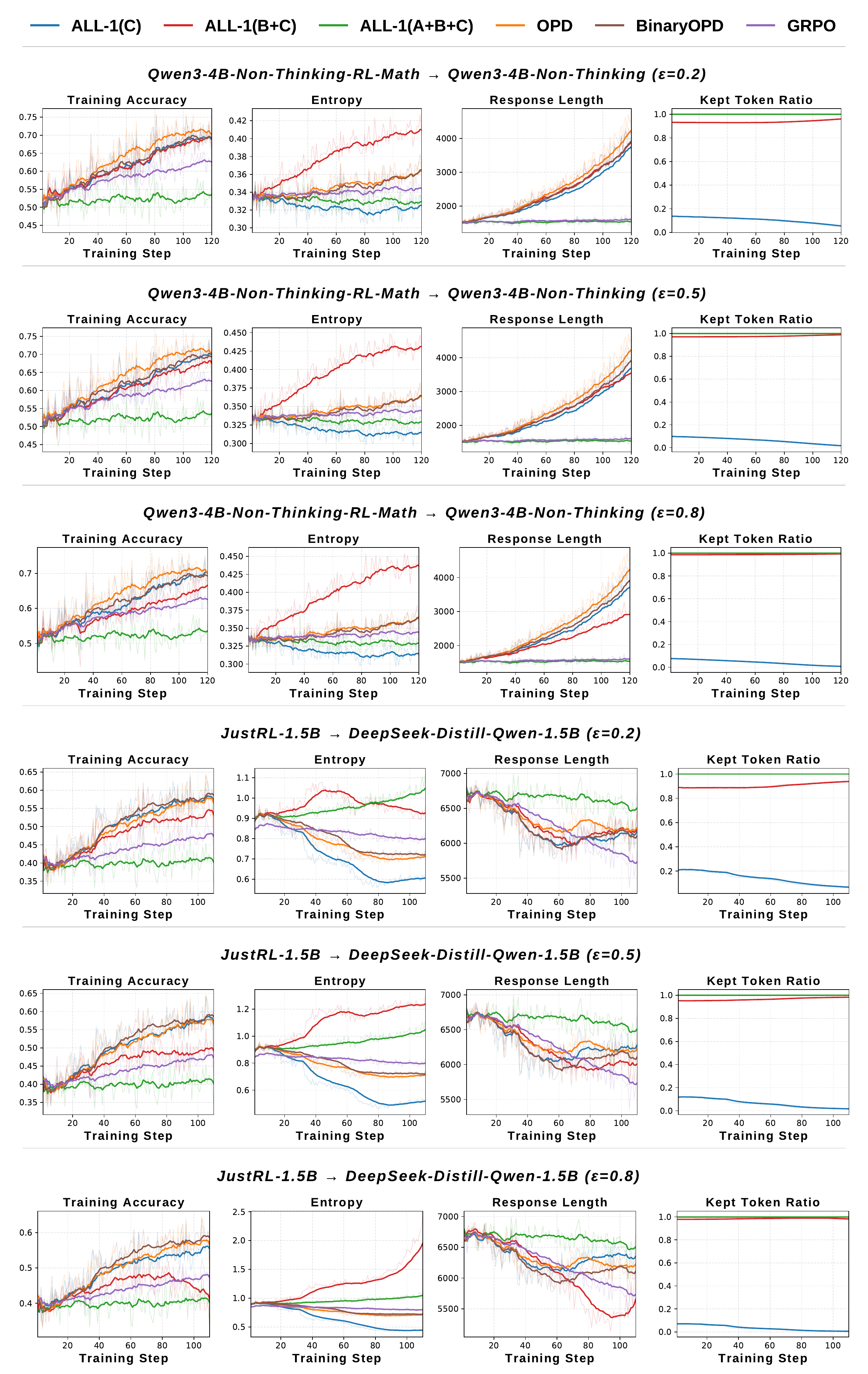}
    \caption{Full training dynamics for the negative reward experiment (ALL-1) across different \(\epsilon\) values and student-teacher pairs.}
    \label{fig:app-negative-dynamics}
\end{figure}

Figures~\ref{fig:app-positive-dynamics} and~\ref{fig:app-negative-dynamics} provide the training dynamics for the positive reward experiment and the negative reward experiment in Section~\ref{sec:high-disagreement}, respectively. In both experiments, successful settings exhibit normal training curves, while settings that incorrectly reward the high-disagreement group fail to converge or degrade clearly.

\section{Visualization of High-Disagreement Tokens}
\label{app:high-disagreement-visualization}

\begin{figure}[h]
    \centering
    \includegraphics[width=\textwidth]{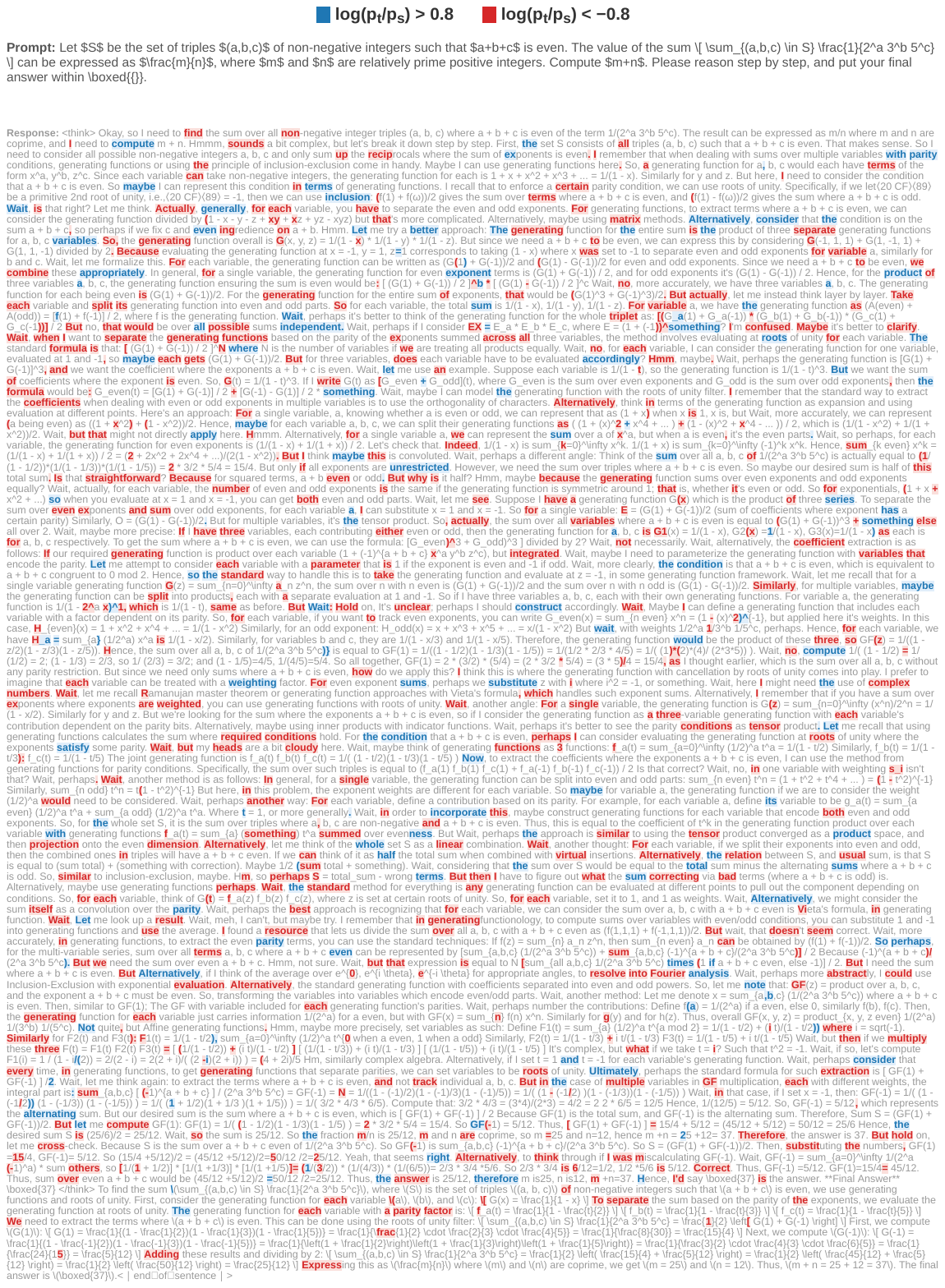}
    \caption{Visualization of high-disagreement tokens. Blue tokens have \(\log(p_T/p_s) > 0.8\) and red tokens have \(\log(p_T/p_s) < -0.8\).}
    \label{fig:token-logratio-example}
\end{figure}

Figure~\ref{fig:token-logratio-example} visualizes an example of high-disagreement tokens. The tokens highlighted in blue have \(\log(p_T/p_s) > 0.8\), and those highlighted in red have \(\log(p_T/p_s) < -0.8\). This example is from the JustRL-1.5B teacher and the DeepSeek-Distill-Qwen-1.5B student.

\section{Hardware Setup}
\label{app:hardware}

All experiments are conducted on NVIDIA RTX 5090, A100, A800, H20, and H100 GPUs.

\end{document}